\documentclass[final]{cvpr}

\usepackage{times}
\usepackage{epsfig}
\usepackage{graphicx}
\usepackage{amsmath}
\usepackage{algorithmicx}
\usepackage{amssymb}
\usepackage{multirow}
\usepackage{color}
\usepackage{enumitem}
\usepackage{xcolor}
\usepackage{bm}
\usepackage{gensymb}
\usepackage[geometry]{ifsym}

\usepackage[pagebackref=true,breaklinks=true,colorlinks,bookmarks=false]{hyperref}

\def\cvprPaperID{10643} 
\def\confYear{CVPR 2021}

\begin{document}

\title{Exploring Complementary Strengths of Invariant and Equivariant Representations for Few-Shot Learning}

\author{Mamshad Nayeem Rizve\IEEEauthorrefmark2
\and Salman Khan\IEEEauthorrefmark3
\and Fahad Shahbaz Khan\IEEEauthorrefmark3
\and Mubarak Shah\IEEEauthorrefmark2\and
{\normalsize \IEEEauthorrefmark2Center for Research in Computer Vision, UCF, USA}\and {\normalsize \IEEEauthorrefmark3Mohamed bin Zayed University of AI, UAE}\and
\texttt{\small nayeemrizve@knights.ucf.edu, \{salman.khan, fahad.khan\}@mbzuai.ac.ae, shah@crcv.ucf.edu}}

\maketitle

\begin{abstract}
In many real-world problems, collecting a large number of labeled samples is infeasible. Few-shot learning (FSL) is the dominant approach to address this issue, where the objective is to quickly adapt to novel categories in presence of a limited number of samples. FSL tasks have been predominantly solved by leveraging the ideas from gradient-based meta-learning and metric learning approaches. However, recent works have demonstrated the significance of powerful feature representations with a simple embedding network that can outperform existing sophisticated FSL algorithms. In this work, we build on this insight and propose a novel training mechanism that simultaneously enforces equivariance and invariance to a general set of geometric transformations. Equivariance or invariance has been employed standalone in the previous works; however, to the best of our knowledge, they have not been used jointly. Simultaneous optimization for both of these contrasting objectives allows the model to jointly learn features that are not only independent of the input transformation but also the features that encode the structure of geometric transformations. These complementary sets of features help generalize well to novel classes with only a few data samples. We achieve additional improvements by incorporating a novel self-supervised distillation objective. Our extensive experimentation shows that even without knowledge distillation our proposed method can outperform current state-of-the-art FSL methods on \emph{five} popular benchmark datasets. Our codes are available at: \url{https://github.com/nayeemrizve/invariance-equivariance}.
 
\end{abstract}
\vspace{-5mm}

\section{Introduction}
In recent years, deep learning methods have made great strides on several challenging problems \cite{he2016deep, szegedy2016rethinking, he2017mask, carreira2017quo, chen2018encoder}. This success can be partially attributed to the availability of large-scale labeled datasets  \cite{imagenet_cvpr09, carreira2017quo, zhou2017places, lin2014microsoft}. However, acquiring large amounts of labeled data is infeasible in several real-world problems due to practical constraints such as the rarity of an event or the high cost of manual annotation. Few-shot learning (FSL) targets this problem by learning a model on a set of base classes and studies its adaptability to novel classes with only a few samples (typically 1-5) 
\cite{pmlr-v70-finn17a, NIPS2016_6385, snell2017prototypical, sung2018learning}. Remarkably, this setting is different from transfer and self/semi-supervised learning that assumes the availability of pretrained models \cite{sharif2014cnn, zamir2018taskonomy, kornblith2019better} or large-amounts of unlabeled data  \cite{doersch2015unsupervised, chen2020simple,NIPS2019_8749_MixMatch}.

\begin{figure}[t]
\centering
  \includegraphics[width=0.95\linewidth]{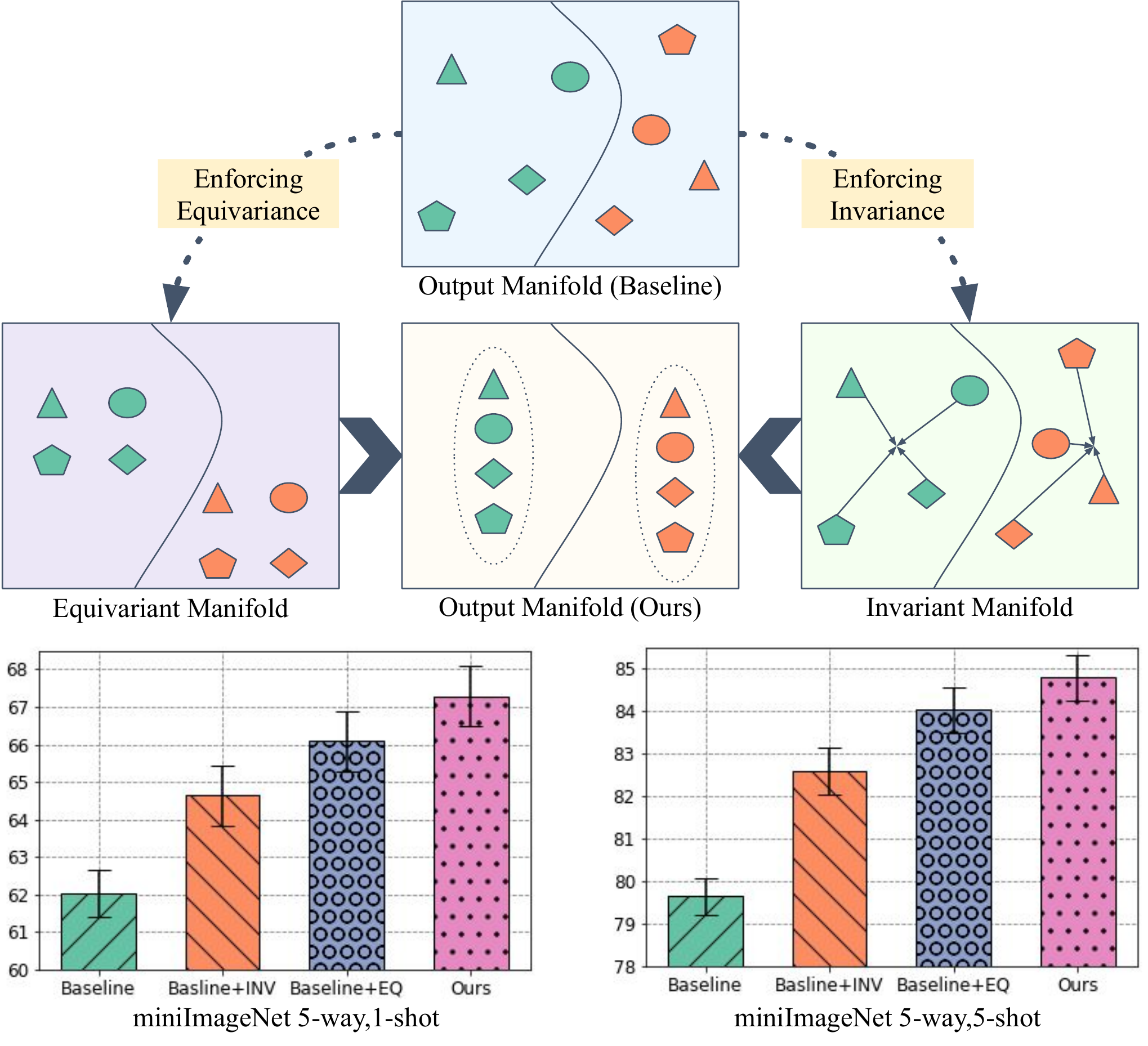}
  \caption{\emph{Approach Overview:} Shapes represent different transformations and colors represent different classes. While the invariant features provide better discrimination, the equivariant features help us learn the internal structure of the data manifold. These complimentary representations help us generalize better to new tasks with only a few training samples. By jointly leveraging the strengths of equivariant and invariant features, our method achieves significant improvement over baseline (bottom row).}
  \vspace{-5mm}
\label{fig:fsl_figure}
\end{figure}

FSL has been predominantly solved using ideas from meta-learning. The two most dominant approaches are optimization-based meta-learning \cite{pmlr-v70-finn17a, jamal2019task, rusu2018metalearning} and metric-learning based methods \cite{snell2017prototypical, sung2018learning, pmlr-v97-allen19b}. Both sets of approaches attempt to train a base learner which can be quickly adapted in the presence of a few novel class examples. However, recently it has been shown in \cite{Raghu2020Rapid} that the quick adaptation of the base learner crucially depends on {\em feature reuse}. Other recent works \cite{tian2020rethink, Dhillon2020A, chen2019closerfewshot} have also shown that a baseline feature extractor trained on all the meta-train set can achieve comparable performance to the state-of-the-art meta learning based methods. This brings in an interesting question: {\em How much  further can FSL performance be pushed by simply improving the base feature extractor?} 

To answer this question, first, we take a look at the inductive biases in machine learning (ML) algorithms. The optimization of all ML algorithms takes advantage of different inductive biases for hypothesis selection; as the solutions are never unique. The generalization of these algorithms often relies on the effective design of inductive biases, since they encode our priori preference for a particular set of solutions. For instance, regularization methods like $\ell_1$/$\ell_2$-penalties \cite{tibshirani1996regression}, dropout \cite{srivastava2014dropout}, or early stopping \cite{prechelt1998early}  implicitly impose Occam's razor in the optimization process by selecting simpler solutions.  
Likewise, convolutional neural networks (CNN) by design impose translation invariance \cite{battaglia2018relational} which makes the internal embeddings translation equivariant. 
Inspired by this, several methods \cite{cohen2016group, finzi2020generalizing, dieleman2016exploiting} have attempted to generalize CNNs by imposing {\em equivariance} to different geometric transformations so that the internal structure of data can be modeled more efficiently. {On the other hand,} methods like \cite{laptev2016ti} try to be robust against nuisance variations by learning transformation {\em invariant} features. However, such inductive biases do not provide optimal generalization on FSL tasks and the design of efficient inductive designs for FSL is relatively unexplored.

In this paper, we propose a novel feature learning approach by designing an effective set of inductive biases. We observe that the features required to achieve invariance against input transformations can provide better discrimination, but do not ensure optimal generalization. Similarly, features that focus on transformation discrimination are not optimal for class discrimination but learn equivariant properties that help in learning the data structure leading to better transferability. Therefore, we propose to combine the complementary strengths of both feature types through a multi-task objective that simultaneously seeks to retain both invariant and equivariant features. We argue that learning such generic features encourages the base feature extractor to be more general. We validate this claim by performing extensive experimentation on multiple benchmark datasets. We also conduct thorough ablation studies to demonstrate that enforcing both equivariance and invariance outperforms enforcing either of these objectives alone (see Fig.~\ref{fig:fsl_figure}). 

Our main contributions are:\vspace{-0.5em}
\begin{itemize}[leftmargin=*]\setlength{\itemsep}{-0.1em}
\item We enforce complimentary {\em equivariance} and {\em invariance} to a general set of geometric transformations to model the underlying structure of the data, while remaining discriminative, thereby improving generalization for FSL.
\item Instead of extensive architectural changes, we propose a simple alternative by defining self-supervised tasks as auxiliary supervision. For {\em equivariance}, we introduce a transformation discrimination task, while an instance discrimination task is developed to learn transformation {\em invariant} features.
\item We demonstrate additional gains with cross-task knowledge distillation that retains the variance properties.

\end{itemize}

\section{Related Works}

 {\bf Few-shot Learning:}  The FSL approaches generally belong to the meta-learning family, which either learn a generalizable metric space \cite{snell2017prototypical,koch2015siamese,vinyals2016matching,oreshkin2018tadam} or apply gradient-based updates to obtain a good initialization. In the first class of methods,  Siamese networks related a pair of images \cite{koch2015siamese}, matching networks applied an LSTM based context encoder to match query and support set images \cite{vinyals2016matching},  and prototypical networks used the distance between the query and the prototype embedding for class assignment \cite{snell2017prototypical}.   A task-dependent metric scaling approach to improve FSL was introduced in \cite{oreshkin2018tadam}. The second category use gradient-based meta-learning methods that include using a sequence model (e.g., LSTM) to learn generalizable optimization rules \cite{ravi2017optimization}, Model-agnostic Meta-Learning (MAML) to find a good initialization that can be quickly adapted to new tasks with minimal supervision \cite{pmlr-v70-finn17a}, and Latent Embedding Optimization (LEO) that applied MAML in the low dimensional space from which high-dimensional parameters can be generated. A few recent efforts, e.g., ProtoMAML \cite{Triantafillou2020Meta-Dataset:}, combined the complementary strengths of metric-learning and gradient-based meta-learning methods.

\textbf{Inductive Biases in CNNs:} Inductive biases reflect our prior knowledge regarding a particular problem. State of the art CNNs are based on such design choices which range from the convolutional operator (e.g., the weight sharing and translational equivariance) \cite{lecun1995convolutional}, pooling operator (e.g., local neighbourhood relevance) \cite{Cohen2017InductiveBO}, regularization mechanisms (e.g., sparsity with $\ell_1$ regularizer) \cite{khan2018guide}, and loss functions (e.g., max-margin boundaries) \cite{hayat2019gaussian}. Similarly, recurrent architectures and attention mechanisms are biased towards preserving contextual information and being invariant to time translation \cite{battaglia2018relational}. A number of approaches have been designed to achieve invariance to nuisances such as natural perturbations \cite{hendrycks2019benchmarking,tramer2019adversarial}, viewpoint changes \cite{milford2015sequence}, and image transformations \cite{cubuk2018autoaugment,buslaev2020albumentations}. On the other hand, equivariant representations have also been investigated to retain knowledge regarding group actions \cite{cohen2016group,qi2020learning,sabour2017dynamic,lenssen2018group}, thereby maintaining meaningful structure in the representations. In this work, we advocate that the representations required to simultaneously achieve invariance and equivariance can be useful for generalization to new tasks with limited data.

\textbf{Self-supervised Learning for FSL:} Our self-supervised loss is inspired by the recent progress in self-supervised learning (SSL), where proxy tasks are defined to learn transferable representations without adding any manual annotations \cite{rajasegaran2020self}. The pretext tasks  include colorization \cite{larsson2016learning,zhang2016colorful}, inpainting \cite{pathak2016context}, relative patch location \cite{doersch2015unsupervised,noroozi2016unsupervised}, and amount of rotation applied \cite{gidaris2018unsupervised}. Recently,   the potential of SSL for FSL was explored in \cite{gidaris2019boosting, su2020does}. In \cite{gidaris2019boosting}
 a parallel branch with the rotation prediction task to help learn generalizable features was added. Su \etal \cite{su2020does} also used rotation and permutation of patches as auxiliary tasks and concluded that SSL is more effective in low-shot regimes and under significant domain shifts. A recent approach employed SimCLR \cite{chen2020simple} style contrastive learning with augmented pairs to learn improved representations in either unsupervised pretraining \cite{medina2020self} or episodic training \cite{doersch2020crosstransformers} for FSL. 

In contrast to the existing SSL approaches for FSL, we propose to jointly optimize for a complimentary pair of pretext tasks that lead to better generalization. Our novel distillation objective acquires knowledge from the classification as well as proxy task heads and demonstrates further performance improvements. We present our approach next.


\section{Our Approach}
\begin{figure*}[t]
\begin{center}
  \includegraphics[width=0.85\linewidth]{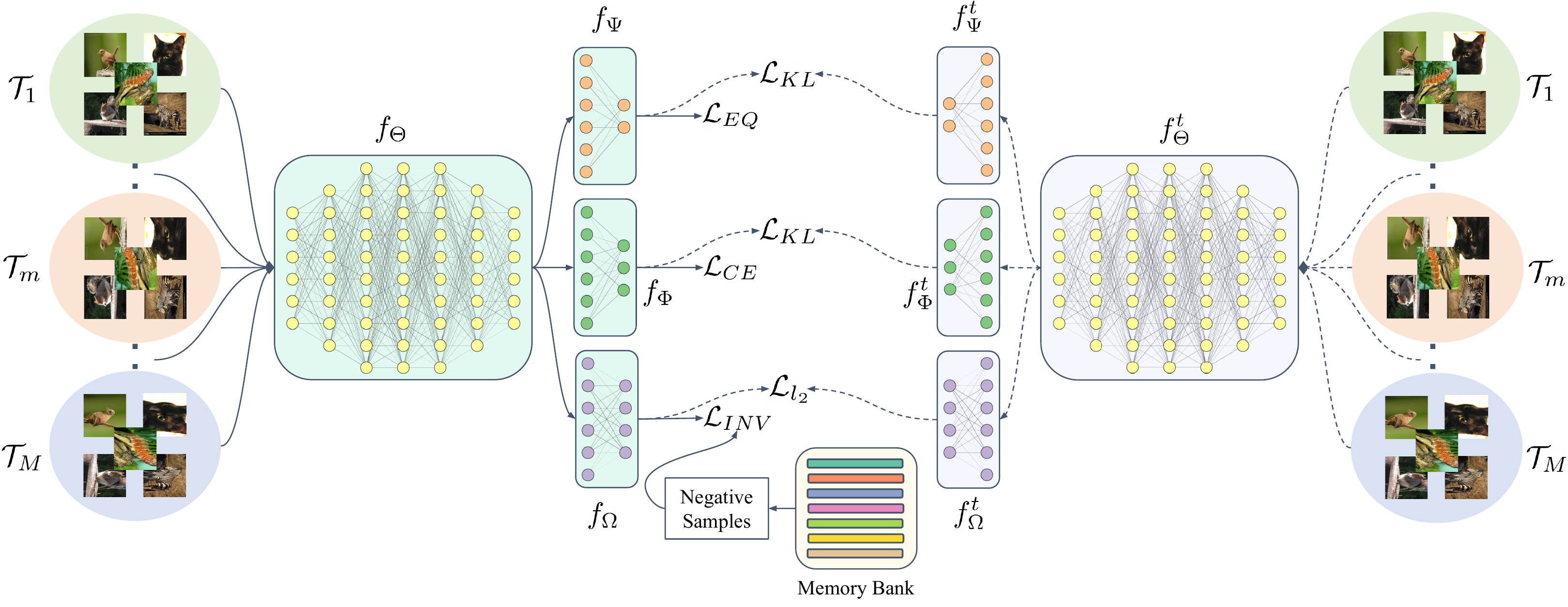}
\end{center}
    \vspace{-4mm}
  \caption{\emph{Network Architecture during Training:} A series of transformed inputs (transformed by applying transformations $\mathcal{T}_1$...$\mathcal{T}_M$) are provided to a shared feature extractor $f_{\Theta}$. The resulting embedding is forwarded to three parallel heads $f_{\Psi}, f_{\Phi}$ and $f_{\Omega}$ that focus on learning equivariant features, discriminative class boundaries, and invariant features, respectively. The resulting output representations are distilled from an old copy of the model (\emph{teacher} model on the right) across multiple-heads to further improve the encoded representations. Notably, a dedicated memory bank of negative samples helps stabilize our invariant contrastive learning.}
  \vspace{-5mm}
\label{fig:arch}
\end{figure*}


We first describe the problem setting and the baseline training approach and then present our proposed approach. 

\subsection{Problem Formulation}
Few-shot learning (FSL) operates in two phases,  first  a model on a set of \emph{base} classes is trained and then at inference a new set of \emph{few-shot} classes are received. We define the base training set as $\mathcal{D}_b = \{(\mathbf{x}, \mathbf{y})\}$,  where $\mathbf{x} \in \mathcal{I} \subset \mathbb{R}^{h\times w\times 3}$ is an image, and the one-hot encoded label $\mathbf{y} \in \mathcal{Y} \subset \mathbb{R}^{N_b}$ can belong to a total of $N_b$ base classes. At inference, a data set of few-shot classes $\mathcal{D}_f = \{(\mathbf{x}, \mathbf{y})\}$ is presented for learning such that the label $\mathbf{y}$ belongs to one of the $N_f$ novel classes, each with a total of $K$ examples ($K$ typically ranges between 1-5). The evaluation setting for few-shot classes is denoted as $N_f$-way, $K$-shot.  Importantly, the $N_b$ base and $N_f$ few-shot classes belong to totally disjoint sets.

For solving the FSL task, most meta-learning methods \cite{pmlr-v70-finn17a, snell2017prototypical, NIPS2016_6385} have leveraged an episodic training scheme. An episode consists of a small train and test set $\left(\mathcal{D}_{tr}^{i}, \mathcal{D}_{ts}^{i}\right)$. The examples for the train and test set of an episode are sampled from the same distribution i.e. from the same subset of meta-training classes. Meta-learning methods try to optimize the parameters of the base learner by solving a collection of these episodes. The main motivation is that the evaluation conditions should be emulated in the base training stage. However, following recent works \cite{tian2020rethink, Dhillon2020A, chen2019closerfewshot}, we do not use an episodic training scheme which allows us to train a single generalizable model that can be efficiently used for any-way, any-shot setting without retraining. Specifically, we train our base learner on the whole base training set  $\mathcal{D}_{b}$ in a supervised manner.


Let's assume our base learner for the FSL task is a neural network, $f_\Theta$, parameterized with parameters $\Theta$. The role of this base learner is to extract good feature embeddings that can generalize for novel classes. The base learner $f_\Theta$ can project an input image $\mathbf{x}$ into the embedding space $f_\Theta: \mathbf{x} \rightarrow \mathbf{z}$, such that $\mathbf{z}\in\mathbb{R}^d$. Now, to optimize the parameters of the base learner $f_\Theta$ we need a classifier to project the extracted embeddings into the label space. To this end, we introduce a classifier function, $f_\Phi$, with parameters $\Phi$ that  projects the embeddings $\mathbf{z}$ into the label space $\mathcal{Y}$ i.e., $f_\Phi: \mathbf{z} \rightarrow \mathbf{\Tilde{y}}$, such that $\mathbf{\Tilde{y}}\in\mathcal{Y}$.

We jointly optimize the parameters of both $f_\Theta$ and $f_\Phi$ by minimizing cross-entropy loss on the whole base-training set $\mathcal{D}_{b}$. The classification loss is given by, 
\begin{align}
   \mathcal{L}_{ce} = -\log \frac{\exp(\mathbf{\Tilde{y}}_{j: \mathbf{y}_j = 1} )}{\sum_i \exp(\mathbf{\Tilde{y}}_i)}, s.t., \mathbf{y} \in \{0,1\}^{N_b}, 
   \mathbf{\Tilde{y}} = f_{\Theta,\Phi}(\mathbf{x}). \notag
\end{align}
To regularize the parameters of both of the sub-networks, we add a regularization term. Hence, the learning objective for our baseline training algorithm becomes:
\begin{equation}
\label{eqn:baseline}
\small
    \mathcal{L}_{baseline} = \underset{(\mathbf{x},\mathbf{y})\sim \mathcal{D}_{b}}{{\mathbb{E}}}\bigg[ \mathcal{L}_{ce}\left(f_{\Theta,\Phi}(\mathbf{x}), \mathbf{y}\right)\bigg] + \mathcal{R}(\Theta,\Phi).
\end{equation}
Here, $\mathcal{R}(\Theta,\Phi)$ is an $\mathcal{L}_2$ regularization term for the parameters $\Theta$ and $\Phi$. Next, we present our inductive objectives.

\subsection{Injecting Inductive Biases through SSL}
 
We propose to enforce equivariance and invariance to a general set of geometric transformations $\mathcal{T}$ by simply performing  self-supervised learning (SSL). Self-supervision is particularly useful for learning general features without accessing semantic labels. For representation learning, self-supervised methods generally aim for either achieving equivariance to some input transformations or learn to discriminate instances by making the representations invariant. To the best of our knowledge, simultaneous equivariance and invariance to a general set of geometric transformations $\mathcal{T}$ has not been explored in the self-supervised literature. We are the first ones to do so.

The transformation set $\mathcal{T}$ can be obtained from a family of geometric transformations, $\mathcal{D}_{\mathcal{T}}$; $\mathcal{T}\sim\mathcal{D}_{\mathcal{T}}$.
%
%
Here, $\mathcal{D}_{\mathcal{T}}$ can be interpreted as a family of geometric transformations like Euclidean transformation, Similarity transformation, Affine transformation, and Projective transformation. All of these geometric transformations can be represented with a $\mathcal{R}^{3\times3}$ matrix with varying degrees of freedom. However, enforcing equivariance and invariance for a continuous space of geometric transformations, $\mathcal{T}$, is difficult and may even lead to suboptimal solutions. To overcome this issue, in this work, we quantize the \emph{complete} space of affine transformations. We approximate $\mathcal{D}_{\mathcal{T}}$ by dividing it into $M$ discrete set of transformations. Here, $M$ can be selected based on the nature of the data and computation budget.

For training, we generate $M$ transformed copies of an input image $\mathbf{x}$ by applying all $M$ transformations. Then we combine all of these transformed images together into a single tensor, $\mathbf{x}_{all} = \{\mathbf{x}_0, \mathbf{x}_1,...,\mathbf{x}_{M-1}\}$. Here, $\mathbf{x}_i$ is the input image $\mathbf{x}$ transformed through $i^{th}$ transformation, $\mathcal{T}_i$ (the subscript of $\mathbf{x}_i$ is dropped in the subsequent discussion for clarity). We send this composite input to the network and optimize for both equivariance and invariance. The training is performed in a multi-task fashion. In addition to the classification head, which is needed for the baseline supervised training, two other heads are added on top of the base learner, as shown in Figure \ref{fig:arch}. One of these heads is used for enforcing equivariance, and the other is used for enforcing invariance. This multi-task training scheme ensures that the base learner retains both transformation equivariant and invariant features in the output embedding. 
We explain each component of our inductive loss below.
 
\vspace{-4mm}

\subsubsection{Enforcing Equivariance}

As discussed above, equivariant features help us encode the inherent structure of data that improves generalization of features to new tasks.  To enforce equivariance for the set $\mathcal{T}$ comprising of $M$ quantized transformations, we introduce an MLP $f_\Psi$ with parameters $\Psi$. The role of $f_\Psi$ is to project the output embeddings from the base learner $\mathbf{z}$ into an equivariant space  i.e., $f_\Psi: \mathbf{z} \rightarrow \mathbf{\Tilde{u}}$, where $\mathbf{\Tilde{u}}\in\mathcal{U}\subset\mathbb{R}^M$.

In order to train the network, we create proxy labels  without any manual supervision. For a specific transformation, a $M$ dimensional one-hot encoded vector $\mathbf{u} \in \{0,1\}^M$ (such that $\sum_i\mathbf{u}_i = 1$) is used to represent the label for $f_\Psi$. Once proxy labels are assigned, training is performed in a supervised manner with the cross-entropy loss, as follows: 
\begin{equation}
\label{eqn:equivariant}
\small
    \mathcal{L}_{eq} = -\log \frac{\exp(\mathbf{\Tilde{u}}_{j: \mathbf{u}_j = 1} )}{\sum_i \exp(\mathbf{\Tilde{u}}_i)},  \; s.t., \mathbf{\Tilde{u}} = f_{\Theta,\Psi}(\mathbf{x}).
\end{equation}
This supervised training with proxy labels in the equivariant space $\mathcal{U}$ ensures that the output embedding $\mathbf{z}$ retains transformation equivariant features. 
\vspace{-4mm}

\subsubsection{Enforcing Invariance}
While equivariant representations are important to encode the structure in data, they may not be optimal for class discrimination. This is because the transformations we consider are nuisance variations that do not change the image class, therefore a good feature extractor should also encode representations that are independent of these input variations. To enforce invariance to the set $\mathcal{T}$ consisting of $M$ quantized transformations, we introduce another MLP $f_\Omega$ with parameters $\Omega$. The role of $f_\Omega$ is to project the output embeddings from the base learner $\mathbf{z}$ into an invariant space i.e., $f_\Omega: \mathbf{z} \rightarrow \mathbf{v}$ where $\mathbf{v}\in\mathcal{V}\subset \mathbb{R}^D$ and $D$ is the dimension of the invariant embedding. 

To optimize for invariance we leverage a contrastive loss \cite{hadsell2006dimensionality} for instance discrimination. We enforce invariance by maximizing the similarity between an embedding $\mathbf{v}^m$ corresponding to a transformed image (after undergoing $m^{th}$ transformation $\mathcal{T}_m$), and the reference embedding $\mathbf{v}^0$ (embedding from the original image without applying any transformation $\mathcal{T}$). Importantly, we note that selecting negatives within a batch is not sufficient to obtain discriminant representations \cite{wu2018unsupervised, misra2020self}.
We employ a memory bank in our contrastive loss to sample more \textit{negative samples} without arbitrarily increasing the batch size. Further, the memory bank allows a stable convergence behavior \cite{misra2020self}. Our learning objective is as follows: 
  
%
%
\begin{equation}
    \label{eqn:invariance}
    \small
    \mathcal{L}_{in} = -\frac{1}{M}\sum_{m=0}^{M-1}\log \left(h(\mathbf{v}^r,\mathbf{v}^m)\right)
\begin{cases}
m\neq0\rightarrow\mathbf{v}^r=\mathbf{v}^0\\
m=0\rightarrow\mathbf{v}^r=\mathbf{\Tilde{v}}^0
\end{cases}
\end{equation}
where, $m$ denotes the transformation index, $\mathbf{\Tilde{v}}^0$ represents a previous copy of the reference $\mathbf{{v}}^0$ held in the memory and the function $h(\cdot)$ is defined as, 


\vspace{-2mm}
\begin{equation}
\label{eqn:contrast}
\small
    h(\mathbf{v}^r,\mathbf{v}^m) =  \frac{\exp\left(\frac{{s(\mathbf{v}^r,\mathbf{v}^m)}}{{\tau}}\right)}{\exp\left(\frac{{s(\mathbf{v}^r,\mathbf{v}^m)}}{{\tau}}\right)+\sum\limits_{\mathbf{v}^{'}\in \mathcal{D}_n} \exp\left(\frac{{s(\mathbf{v}^{'},\mathbf{v}^m)}}{{\tau}}\right)}. \notag
\end{equation}

\noindent Here,  $s(.)$ is a similarity function, $\tau$ is the temperature, and $\mathcal{D}_n$ is the set of \textit{negative samples} drawn from the memory bank for a particular minibatch. Note that we also maximize the similarity between the reference embedding $\mathbf{v}^0$ and its past representation $\mathbf{\Tilde{v}}^0$ which helps stabilize the learning. 

\vspace{-2mm}
\subsubsection{Multi-head Distillation}
Once the invariant and equivariant representations are learned by our model, we use self-distillation to train a new model using outputs from the previous model as anchor points (Fig.~\ref{fig:arch}). Note that in typical knowledge distillation \cite{44873}, information is exchanged from a larger model (teacher) to a smaller one (student) by matching their softened outputs. In contrast, the outputs from the same models are matched in the self-distillation \cite{DBLP:conf/icml/FurlanelloLTIA18} where the smooth predictions encode inter-label dependencies, thereby helping the model to learn better representations.

In our case, a simple knowledge distillation by pairing the logits \cite{tian2020rethink} would not ensure the transfer of invariant and equivariant representations learned by the previous model version.
Therefore, we extend the idea of logit-based knowledge distillation and employ it to our invariant and equivariant embedding embeddings. Specifically, in parallel to minimizing the Kullback Leibler (KL) divergence for the soft output of supervised classifier head $f_\Phi$, we also minimize the KL divergence between the outputs of the equivariant head $f_{\Psi}$. Since the output of our invariant head $f_{\Omega}$ is not a probability distribution, we minimize a $\mathcal{L}_2$ loss for distilling the knowledge at this head. The overall learning objective for knowledge distillation is as follows:
\begin{align}
    \label{eqn:knowledge}
    \small
    \mathcal{L}_{\textit{kd}} = & \mathrm{KL}(f_{\Theta,\Phi}^t(\mathbf{x}), f_{\Theta,\Phi}(\mathbf{x})) + \mathrm{KL}(f_{\Theta,\Psi}^t(\mathbf{x}), f_{\Theta,\Psi}(\mathbf{x})) \notag\\
     & + \mathcal{L}_2(f_{\Theta,\Omega}^t(\mathbf{x}), f_{\Theta,\Omega}(\mathbf{x})).
\end{align}
Here, $f_{(.,.)}^t$  and $f_{(.,.)}$  are the teacher and student networks for distillation, respectively.  

\vspace{-2mm}
\subsubsection{Overall Objective}
Finally, we obtain the resultant loss for injecting the desired inductive biases by combining both equivariant $\mathcal{L}_{eq}$, invariant $\mathcal{L}_{in}$, and multi-head distillation $\mathcal{L}_{kd}$ losses:
\begin{align}
\small
   & \mathcal{L}_{inductive} =  \underset{\mathbf{x}\sim \mathcal{D}_{b}, \mathbf{v}' \sim \mathcal{D}_n}{{\mathbb{E}}}\bigg[ \mathcal{L}_{eq} (f_{\Theta,\Psi}(\mathbf{x}), \mathbf{u}) + \notag\\ 
    & \mathcal{L}_{in}(f_{\Theta,\Omega}(\mathbf{x}), \mathbf{v}')  + \mathcal{L}_{kd}(f_{\Theta,\Phi}^{.,t}(\mathbf{x}), f_{\Theta,\Psi}^{.,t}(\mathbf{x}), f_{\Theta,\Omega}^{.,t}(\mathbf{x})) \bigg].\notag
\end{align}
The overall loss is simply a combination of inductive and baseline objectives,
\begin{align}\label{eq:overall}
    \small
    \mathcal{L} =  \mathcal{L}_{baseline} +  \mathcal{L}_{inductive}.
\end{align}

\subsection{Few-Shot Evaluation}
For evaluation, we test our base learner $f_\Theta$ by sampling FSL tasks from a held-out test set comprising of images from novel classes. Each FSL task contains a support set and a corresponding query set \{$D_{supp}$, $D_{query}$\}; both contain images from the same subset of test classes. Using $f_\Theta$, we obtain embeddings for the images of both $D_{supp}$ and $D_{query}$. Following \cite{tian2020rethink}, we train a simple logistic regression classifier based on the image embeddings and the corresponding labels from the $D_{supp}$. We use that linear classifier to infer the labels of the query embeddings.


\section{Experimental Evaluation}

\noindent\textbf{Datasets:}
We evaluate our method on five popular benchmark FSL datasets. Two of these datasets are subset of the CIFAR100 dataset: CIFAR-FS \cite{bertinetto2018metalearning} and FC100 \cite{oreshkin2018tadam}. Another two are derivatives of the ImageNet \cite{imagenet_cvpr09} dataset: miniImageNet \cite{NIPS2016_6385}  and tieredImageNet \cite{ren2018meta}. The CIFAR-FS dataset is constructed by randomly splitting the 100 classes of the CIFAR-100 dataset into 64, 16, and 20 train, validation, and test splits. FC100 dataset makes the FSL task more challenging by making the splits more diverse; the FC100 train, validation, and test splits contain 60, 20, and 20 classes. Following \cite{Ravi2017OptimizationAA}, we use 64, 16, and 20 classes of the miniImageNet dataset for training, validation, and testing. The tieredImageNet dataset contains 608 ImageNet classes that are grouped into 34 high-level categories, and we use 20/351, 6/97, and 8/160 categories/classes for training, validation, and testing. We also evaluate our method on the newly proposed Meta-Dataset \cite{Triantafillou2020Meta-Dataset:}, which contains 10 diverse datasets to make the FSL task more challenging and closer to realistic classification problems.



\begin{table}[h]
\begin{center}
\small
\resizebox{\columnwidth}{!}{%
\begin{tabular}{@{\hskip0pt}l@{\hskip4pt}c@{\hskip8pt}c@{\hskip8pt}c@{\hskip1pt}}
\hline

\hline

\hline\\[-3mm]
\textbf{Methods} & \textbf{Backbone} & \textbf{1-Shot} & \textbf{5-Shot}
\\[-3mm]
\\
\hline

\hline

\hline
MAML\cite{pmlr-v70-finn17a} & 32-32-32-32 & $58.90\pm1.90$ & $71.50\pm1.00$ \\
Proto-Net$^{{\dagger}}$\cite{snell2017prototypical} & 64-64-64-64 & $55.50\pm0.70$ & $72.00\pm0.60$ \\ 
Relation Net\cite{sung2018learning} &  64-96-128-256 & $55.00\pm1.00$ & $69.30 \pm0.80$ \\
R2D2\cite{bertinetto2018metalearning} & 96-192-384-512 & $65.30\pm0.20$ & $79.40\pm0.10$ \\
Shot-Free\cite{ravichandran2019few} & ResNet-12 & $69.20$ & $84.70$ \\
TEWAM\cite{qiao2019transductive} & ResNet-12 & $ 70.40$ & $81.30$ \\
Proto-Net$^{{\dagger}}$\cite{snell2017prototypical} & ResNet-12 & $72.20\pm0.70$ & $83.50\pm0.50$ \\ 
MetaOptNet\cite{lee2019meta} & ResNet-12 & $72.60\pm0.70$ & $84.30\pm0.50$ \\
Boosting\cite{gidaris2019boosting} & WRN-28-10 & $73.60\pm0.30$ & $86.00\pm0.20$ \\
Fine-tuning\cite{Dhillon2020A} & WRN-28-10 & \textcolor{blue}{$76.58\pm0.68$} & $85.79\pm0.50$  \\
DSN-MR\cite{simon2020adaptive} & ResNet-12 & $75.60\pm0.90$ & $86.20\pm0.60$ \\
MABAS\cite{10.1007/978-3-030-58452-8_35} & ResNet-12 & $73.51\pm0.92$ & $85.49\pm0.68$ \\
RFS-Simple\cite{tian2020rethink} & ResNet-12 & $71.50\pm0.80$ & $86.00\pm0.50$ \\
RFS-Distill\cite{tian2020rethink} & ResNet-12 & $73.90\pm0.80$ & \textcolor{blue}{$86.90\pm0.50$} \\

\hline
Ours & ResNet-12 & $76.83\pm0.82$ & $89.26\pm0.58$ \\
Ours-Distill & ResNet-12 & \textcolor{red}{$77.87\pm0.85$} & \textcolor{red}{$89.74\pm0.57$} \\ \hline

\hline

\hline
\end{tabular}}
\end{center}
\vspace{-2mm}
\caption{Average 5-way few-shot classification accuracy with 95\% confidence intervals on {\bf CIFAR-FS} dataset; ${^\dagger}$trained on train and validation sets. Top two results are shown in red and blue.}
\label{tab:cifarfs}
\vspace{-3mm}
\end{table}

\begin{table}[ht!]
\begin{center}
\small
\resizebox{\columnwidth}{!}{%
\begin{tabular}{@{\hskip0pt}l@{\hskip10pt}c@{\hskip10pt}c@{\hskip10pt}c@{\hskip1pt}}
\hline

\hline

\hline\\[-3mm]
\textbf{Methods} & \textbf{Backbone} & \textbf{1-Shot} & \textbf{5-Shot}
\\[-3mm]
\\
\hline

\hline

\hline
Proto-Net$^{{\dagger}}$\cite{snell2017prototypical} & 64-64-64-64 & $35.30\pm0.60$ & $48.60\pm0.60$ \\ 
Proto-Net$^{{\dagger}}$\cite{snell2017prototypical} & ResNet-12 & $37.50\pm0.60$ & $52.50\pm0.60$ \\ 
TADAM\cite{oreshkin2018tadam} & ResNet-12 & $40.10\pm0.40$ & $56.10\pm0.40$ \\
MetaOptNet\cite{lee2019meta} & ResNet-12 & $41.10\pm0.60$ & $55.50\pm0.60$ \\
MTL\cite{sun2019meta} & ResNet-12 & \textcolor{blue}{$45.10\pm1.80$} & $57.60\pm0.90$ \\
Fine-tuning\cite{Dhillon2020A} & WRN-28-10 & $43.16\pm0.59$ & $57.57\pm0.55$  \\
MABAS\cite{10.1007/978-3-030-58452-8_35} & ResNet-12 & $42.31\pm0.75$ & $57.56\pm0.78$ \\
RFS-Simple\cite{tian2020rethink} & ResNet-12 & $42.60\pm0.70$ & $59.10\pm0.60$ \\
RFS-Distill\cite{tian2020rethink} & ResNet-12 & $44.60\pm0.70$ & \textcolor{blue}{$60.90\pm0.60$} \\
\hline
Ours & ResNet-12 & $47.38\pm0.79$ & $64.43\pm0.77$ \\
Ours-Distill & ResNet-12 & \textcolor{red}{$47.76\pm0.77$} & \textcolor{red}{$65.30\pm0.76$} \\ \hline

\hline

\hline
\end{tabular}}
\end{center}
\vspace{-2mm}
\caption{Average 5-way few-shot classification accuracy with 95\% confidence intervals on {\bf FC100} dataset; ${^\dagger}$trained on train and validation sets. Top two results are shown in red and blue.}
\label{tab:fc100}
\vspace{-5mm}
\end{table}

\begin{table}[t]
\begin{center}
\small
\resizebox{\columnwidth}{!}{%
\begin{tabular}{@{\hskip0pt}l@{\hskip1pt}c@{\hskip5pt}c@{\hskip8pt}c@{\hskip0pt}}
\hline

\hline

\hline\\[-3mm]
\textbf{Methods} & \textbf{Backbone} & \textbf{1-Shot} & \textbf{5-Shot}
\\[-3mm]
\\
\hline

\hline

\hline
MAML\cite{pmlr-v70-finn17a} & 32-32-32-32 & $48.70\pm1.84$ & $63.11\pm0.92$ \\
Matching Net \cite{NIPS2016_6385} & 64-64-64-64 &  $43.56\pm0.84$ &  $55.31\pm0.73$ \\ 
Proto-Net$^{{\dagger}}$\cite{snell2017prototypical} & 64-64-64-64 & $49.42\pm0.78$ & $68.20\pm0.66$ \\ 
Relation Net\cite{sung2018learning} &  64-96-128-256 & $50.44\pm0.82$ & $65.32\pm0.70$ \\
R2D2\cite{bertinetto2018metalearning} & 96-192-384-512 & $51.20\pm0.60$ & $68.80\pm0.10$ \\
SNAIL\cite{mishra2018a} & ResNet-12 & $55.71\pm0.99$ & $68.88\pm0.92$ \\
AdaResNet\cite{pmlr-v80-munkhdalai18a} & ResNet-12 & $56.88\pm0.62$ & $71.94\pm0.57$ \\
TADAM\cite{oreshkin2018tadam} & ResNet-12 & $58.50\pm0.30$ & $76.70\pm0.30$ \\
Shot-Free\cite{ravichandran2019few} & ResNet-12 & $59.04$ & $77.64$ \\
TEWAM\cite{qiao2019transductive} & ResNet-12 & $60.07$ & $75.90$ \\
MTL\cite{sun2019meta} & ResNet-12 & $61.20\pm1.80$ & $75.50\pm0.80$ \\
MetaOptNet\cite{lee2019meta} & ResNet-12 & $62.64\pm0.61$ & $78.63\pm0.46$ \\
Boosting\cite{gidaris2019boosting} & WRN-28-10 & $63.77\pm0.45$ & $80.70\pm0.33$ \\
Fine-tuning\cite{Dhillon2020A} & WRN-28-10 & $57.73\pm0.62$ & $78.17\pm0.49$ \\
LEO-trainval$^{{\dagger}}$\cite{rusu2018metalearning} & WRN-28-10 & $61.76\pm0.08$ & $77.59\pm0.12$ \\
Deep DTN\cite{Chen2020DiversityTN} & ResNet-12 & $63.45\pm0.86$ & $77.91\pm0.62$ \\
AFHN\cite{li2020adversarial} & ResNet-18 & $62.38\pm0.72$ & $78.16\pm0.56$ \\
AWGIM\cite{guo2020attentive} & WRN-28-10 & $63.12\pm0.08$ & $78.40\pm0.11$ \\
DSN-MR\cite{simon2020adaptive} & ResNet-12 & $64.60\pm0.72$ & $79.51\pm0.50$ \\
MABAS\cite{10.1007/978-3-030-58452-8_35} & ResNet-12 & \textcolor{blue}{$65.08\pm0.86$} & \textcolor{blue}{$82.70\pm0.54$} \\
RFS-Simple\cite{tian2020rethink} & ResNet-12 & $62.02\pm0.63$ & $79.64\pm0.44$ \\
RFS-Distill\cite{tian2020rethink} & ResNet-12 & $64.82\pm0.60$ & $82.14\pm0.43$ \\
\hline
Ours & ResNet-12 & $66.82\pm0.80$ & $84.35\pm0.51$ \\
Ours-Distill & ResNet-12 & \textcolor{red}{$67.28\pm0.80$} & \textcolor{red}{$84.78\pm0.52$} \\ \hline

\hline

\hline
\end{tabular}}
\end{center}
\vspace{-2mm}
\caption{Average 5-way few-shot classification accuracy with 95\% confidence intervals on {\bf miniImageNet} dataset; ${^\dagger}$trained on train and validation sets. Top two results are shown in red and blue.}
\label{tab:minimagenet}
\vspace{-5mm}
\end{table}

\noindent\textbf{Implementation Details:}
Following \cite{tian2020rethink, mishra2018a, oreshkin2018tadam, lee2019meta}, we use a ResNet-12 network as our base learner to conduct experiments on CIFAR-FS, FC100, miniImageNet, tieredImageNet datasets. Following \cite{tian2020rethink, lee2019meta}, we also apply Dropblock \cite{ghiasi2018dropblock} regularizer to our Resnet-12 base learner. For Meta-Dataset experiments we use a Resnet-18 \cite{he2016deep} network as our base learner to be consistent with \cite{tian2020rethink}. We instantiate both of our equivariant and invariant embedding learners ($f_\Psi$, $f_\Omega$) with an MLP consisting of a single hidden layer. The classifier, $f_\Phi$, is instantiated with a single linear layer.

We use SGD optimizer with a momentum of 0.9 in all experiments. For CIFAR-FS, FC100, miniImageNet, tieredImageNet datasets we set the initial learning rate to 0.05 and use a weight decay of $5e-4$. For experiments on CIFAR-FS, FC100, miniImageNet datasets, we train for 65 epochs; the learning rate is decayed by a factor of 10 after the first 60 epochs. We train for 60 epochs for experiments on the tieredImageNet dataset; the learning rate is decayed by a factor of 10 for 3 times after the first 30 epochs. For Meta-Dataset experiments, we set the initial learning rate to 0.1 and use a weight decay of $1e-4$. We train our method for 90 epochs and decay the learning rate by a factor of 10 every 30 epochs. We use a  batch size of 64 in all of our experiments except on Meta-Dataset where the batch size is set to 256 following \cite{tian2020rethink}. For Meta-dataset experiments, we use standard data augmentation which includes random horizontal flip and random resized crop. For all the other dataset experiments we use random crop, color jittering and random horizontal flip for data augmentation following \cite{tian2020rethink, lee2019meta}. Consistent with \cite{tian2020rethink}, we use a temperature coefficient of 4.0 for our knowledge distillation experiments. For all datasets, we perform one stage of distillation. We sample 600 FSL tasks to report our scores on all datasets except Meta-Dataset.   

For our geometric transformations, we sample from a complete space of similarity transformation and use four rotation transformations: \{0\degree, 90\degree, 180\degree, 270\degree\}, two scaling transformations: \{0.67, 1.0\} and three aspect ration transformations: \{0.67, 1.0, 1.33\}. These geometric transformations can be applied in any combination. For all of our experiments, we set the total number of applied transformations to 16. Additional details and experiments with more geometric transformations are included in the supplementary materials. For the contrastive loss, we use a memory bank that stores 64-dimensional embedding of instances; we sample 6400 negative samples from the memory bank for each mini-batch and set the value of $\tau$ to 1.0.

\subsection{Results}

\begin{table}[t]
\begin{center}
\small
\resizebox{\columnwidth}{!}{%
\begin{tabular}{@{\hskip0pt}l@{\hskip1pt}c@{\hskip6pt}c@{\hskip8pt}c@{\hskip1pt}}
\hline

\hline

\hline\\[-3mm]
\textbf{Methods} & \textbf{Backbone} & \textbf{1-Shot} & \textbf{5-Shot}
\\[-3mm]
\\
\hline

\hline

\hline
MAML\cite{pmlr-v70-finn17a} & 32-32-32-32 & $51.67\pm1.81$ & $70.30\pm 1.75$ \\
Proto-Net$^{{\dagger}}$\cite{snell2017prototypical} & 64-64-64-64 & $53.31\pm0.89$ & $72.69\pm0.74$ \\ 
Relation Net\cite{sung2018learning} &  64-96-128-256 & $54.48\pm0.93$ & $71.32\pm0.78$ \\
Shot-Free\cite{ravichandran2019few} & ResNet-12 & $63.52$ & $82.59$ \\
MetaOptNet\cite{lee2019meta} & ResNet-12 & $65.99\pm0.72$ & $81.56\pm0.53$ \\
Boosting\cite{gidaris2019boosting} & WRN-28-10 & $70.53\pm0.51$ & $84.98\pm0.36$ \\
Fine-tuning\cite{Dhillon2020A} & WRN-28-10 & $66.58\pm0.70$ & $85.55\pm 0.48$ \\
LEO-trainval$^{{\dagger}}$\cite{rusu2018metalearning} & WRN-28-10 & $66.33\pm0.05$ & $81.44\pm0.09$ \\
AWGIM\cite{guo2020attentive} & WRN-28-10 & $67.69\pm0.11$ & $82.82\pm 0.13$ \\
DSN-MR\cite{simon2020adaptive} & ResNet-12 & $67.39\pm0.82$ & $82.85\pm0.56$ \\
RFS-Simple\cite{tian2020rethink} & ResNet-12 & $69.74\pm0.72$ & $84.41\pm0.55$ \\
RFS-Distill\cite{tian2020rethink} & ResNet-12 & \textcolor{blue}{$71.52\pm0.69$} & \textcolor{blue}{$86.03\pm0.49$} \\

\hline
Ours & ResNet-12 & $71.87\pm0.89$ & $86.82\pm0.58$ \\
Ours-Distill & ResNet-12 & \textcolor{red}{$72.21\pm0.90$} & \textcolor{red}{$87.08\pm0.58$} \\ \hline

\hline

\hline
\end{tabular}}
\end{center}
\vspace{-2mm}
\caption{Average 5-way few-shot classification accuracy with 95\% confidence intervals on {\bf tieredImageNet} dataset; ${^\dagger}$trained on train and validation sets. Top two results are shown in red and blue.}
\label{tab:tiredimagenet}
\vspace{-2mm}
\end{table}

We present our results on five popular benchmark FSL datasets in Table \ref{tab:cifarfs}-\ref{tab:meta_dataset} which demonstrates that even without multi-head distillation our proposed method consistently outperforms the current state-of-the-art (SOTA) FSL methods on both 5-way 1-shot and 5-way 5-shot tasks. By virtue of our novel representation learning approach which retains both the transformation invariant and equivariant features in the learned embeddings, our proposed method improves over the baseline RFS-Simple \cite{tian2020rethink} method across all datasets by 2-5\% for both 1-shot and 5-shot tasks. To be more specific, our method outperforms the current best results on CIFAR-FS dataset (Table \ref{tab:cifarfs}) by 1.3\% in the 1-shot task whereas for the 5-shot task it improves the score by 2.8\%. However, unlike \cite{Dhillon2020A}, which achieves the current best results on the CIFAR-FS 1-shot task, we do not perform any transductive fine-tuning. For FC100 dataset (Table \ref{tab:fc100}) we observe an even bigger improvement; 2.7\% and 4.4\% for 1 and 5-shot, respectively. We see similar trends in miniImageNet and tieredImageNet (Table \ref{tab:minimagenet},\ref{tab:tiredimagenet}) where we consistently improve over the current SOTA methods by 0.7-2.2\%.    

For the Meta-Dataset \cite{Triantafillou2020Meta-Dataset:}, we train our model on the ILSVRC train split and test on  10 diverse datasets. Our results in Table \ref{tab:meta_dataset} demonstrate that our method outperforms the fo-Proto-MAML \cite{Triantafillou2020Meta-Dataset:} across all 10 datasets. Even without multi-head distillation, we outperform both simple and distilled version of the RFS method on 6 out of 10 datasets. 
Overall, we perform favorably well against the RFS, achieving a new SOTA result on the challenging Meta-Dataset.

\begin{table}
\begin{center}
\small
\resizebox{\columnwidth}{!}{%
\begin{tabular}{@{\hskip0pt}lccccc@{\hskip0pt}}
\hline

\hline

\hline\\[-3mm]
 \multicolumn{1}{l}{\multirow{2}{*}{\textbf{Dataset}}} &
 \multicolumn{1}{c}{\textbf{fo-Proto-}} &
 \multicolumn{2}{c}{\textbf{RFS}} &
 \multicolumn{1}{c}{\multirow{2}{*}{\textbf{Ours}}} & 
 \multicolumn{1}{c}{\multirow{2}{*}{\textbf{Ours-Distill}}}
 \\
 &\textbf{MAML} & Simple  & Distill & &
 \\[-3mm]
\\
 \hline

\hline

\hline
ILSVRC & $49.53$ & $60.14$ & \textcolor{red}{$61.48$} & $60.64$ & \textcolor{blue}{$61.36$} \\
Omniglot & $63.37$ & \textcolor{blue}{$64.92$} & $64.31$ & \textcolor{red}{$65.55$} & $65.53$\\
Aircraft & $55.95$ & \textcolor{blue}{$63.12$} & $62.32$ & $65.65$ & \textcolor{red}{$66.58$}\\
Birds & $68.66$ & $77.69$ & \textcolor{red}{$79.47$} & $77.84$ & \textcolor{blue}{$78.23$}\\ 
Textures & $66.49$ & $78.59$ & \textcolor{blue}{$79.28$} & \textcolor{red}{$81.07$} & $80.42$\\ 
Quick Draw & $51.52$ & \textcolor{red}{$62.48$} & $60.83$ & $57.91$ & \textcolor{blue}{$59.02$}\\ 
Fungi & $39.96$ & $47.12$ & \textcolor{blue}{$48.53$} & $49.26$ & \textcolor{red}{$49.50$}\\ 
VGG Flower & $87.15$ & \textcolor{blue}{$91.60$} & $91.00$ & $92.06$ & \textcolor{red}{$92.66$}\\ 
Traffic Signs & $48.83$ & \textcolor{blue}{$77.51$} & $76.33$ & $78.92$ & \textcolor{red}{$79.92$}\\ 
MSCOCO & $43.74$ & $57.00$ & \textcolor{red}{$59.28$} & $55.07$ & \textcolor{blue}{$55.68$}\\\hline
Mean Accuracy & $57.52$ & $68.02$ & \textcolor{blue}{$68.28$} & $68.40$ & \textcolor{red}{$68.89$}\\\hline 

\hline

\hline
\end{tabular}}
\end{center}
\vspace{-2mm}
\caption{Results on Meta-Dataset. Average accuracy (\%) is reported with variable number of ways and shots, following the setup in \cite{Triantafillou2020Meta-Dataset:}. 1000 tasks are sampled for evaluation. Top two results are shown in red and blue.}
\label{tab:meta_dataset}
\vspace{-5mm}
\end{table}


\subsection{Ablations}


\begin{table*}[ht]
\begin{center}
\small
\resizebox{\textwidth}{!}{%
\begin{tabular}{c@{\hskip8pt}l@{\hskip8pt}c@{\hskip8pt}c@{\hskip8pt}c@{\hskip8pt}c@{\hskip8pt}c}
\hline

\hline

\hline\\[-3mm]
 \multicolumn{1}{c}{\multirow{2}{*}{\textbf{Method}}} &
 \multicolumn{2}{c}{\textbf{miniImageNet, 5-Way}} &
 \multicolumn{2}{c}{\textbf{CIFAR-FS, 5-Way}} & \multicolumn{2}{c}{\textbf{FC100, 5-Way}} \\  
\multicolumn{1}{c}{} & \textbf{1-Shot} & \textbf{5-Shot} & \textbf{1-Shot} & \textbf{5-Shot} & \textbf{1-Shot} & \textbf{5-Shot}
 \\[-3mm]
\\
 \hline

\hline

\hline
Baseline Training & $62.02\pm0.63$ & $79.64\pm0.44$ & $71.50\pm0.80$ & $86.00\pm0.50$ & $42.60\pm0.70$ & $59.10\pm0.60$ \\
Ours with only Invariance & $64.64\pm0.80$ & $82.59\pm0.54$ & $73.50\pm0.86$ & $87.55\pm0.61$ & $46.10\pm0.78$ & $63.18\pm0.76$ \\
Ours with only Equivariance & $66.09\pm0.80$ & $84.03\pm0.53$ & $76.37\pm0.83$ & $89.08\pm0.58$ & $46.73\pm0.79$ & $64.09\pm0.75$\\
Ours with Equi and Invar (W/O KD) & $66.82\pm 0.80$ & $84.35\pm 0.51$ & $76.83\pm0.82$ & $89.26\pm0.58$ & $47.38\pm0.79$ & $64.43\pm0.77$ \\
Ours with Supervised KD & $66.95\pm0.78$ & $84.39\pm0.52$ & $76.92\pm0.85$ & $89.34\pm0.57$ & $47.70\pm0.81$ & $65.09\pm0.76$ \\ 
Ours Full & $67.28\pm0.80$ & $84.78\pm0.52$ & $77.87\pm0.85$ & $89.74\pm0.57$ & $47.76\pm0.77$ & $65.30\pm0.76$\\ \hline 

\hline

\hline
\end{tabular}}
\end{center}
\vspace{-2mm}
\caption{Ablation study on \textbf{miniImageNet}, \textbf{CIFAR-FS}, and \textbf{FC100} datasets.}
\label{tab:ablation}
\vspace{-3mm}
\end{table*}

To study the contribution of different components of our method we do a thorough ablation study on three benchmark FSL datasets: miniImageNet, CIFAR-FS, and FC100 (Table \ref{tab:ablation}). On these three datasets, our baseline supervised training achieves 62.02\%, 71.50\%, and 42.60\% average accuracy on 5-way 1-shot task respectively; which is the same as RFS-Simple~\cite{tian2020rethink}. By enforcing invariance we obtain 2.62\%, 2\%, and 3.5\% improvements respectively. Likewise, enforcing equivariance gives 4.07\%, 4.87\%, and 4.13\% improvements over the baseline respectively. On the other hand, we get even bigger improvements by simultaneously optimizing for both equivariance and invariance; achieving 4.8\%, 5.33\%, and 4.78\% improvements on top of the baseline supervised training. Besides, joint training gives 1.3\%-3.3\% improvement over only invariance training and 0.5\%-0.7\% improvement in comparison to only equivariance training. We observe similar trends for 5-way 5-shot task. This consistent improvement across all datasets for both tasks empirically validates our claim that joint optimization for both equivariance and invariance is beneficial for FSL tasks. Our ablation study also shows that the multi-head distillation improves the performance over the standard logit-level supervised distillation across all datasets.

\noindent{\bf Effect of the number of Transformations:} To investigate the effect of the total number of applied transformations, we perform an ablation study on the CIFAR-FS validation set by varying the number of transformations, $M$. We present the results in Table \ref{tab:m_ablation}, which demonstrates that initially, the performance of our method improves with the increasing $M$. However, the performance starts to saturate beyond a particular point. We hypothesize that the performance for an increasing number of transformations decreases since discriminating a higher number of transformations is more difficult and the model spends more representation capability for solving this harder task. A similar trend is observed in \cite{gidaris2018unsupervised}, where increasing the number of recognizable rotations does not lead to better performance. Based on Table \ref{tab:m_ablation} results, we set the value of $M$ to 16 for all of our experiments and do not {\em tune} the $M$ value from dataset to dataset.








\begin{table}
\begin{center}
\small
\begin{tabular}{@{\hskip0pt}c@{\hskip5pt}c@{\hskip6pt}c@{\hskip8pt}c@{\hskip1pt}}
\hline

\hline

\hline\\[-3mm]
 $M$ & \textbf{Description}  & \textbf{1-Shot} & \textbf{5-Shot} 
 \\[-3mm]
\\
 \hline

\hline

\hline
3 & Aspect-Ratio & $65.13\pm0.93$ & $81.22\pm0.66$ \\
4 & Rotation & $66.56\pm0.92$ & $82.64\pm0.64$ \\ 
8 & Rotation, Scale & $67.46\pm0.92$ & $82.80\pm0.64$ \\ 
12 & Aspect-Ratio, Rotation & $68.04\pm0.93$ & $83.48\pm0.64$ \\ 
16 & Aspect-Ratio, Rotation, Scale & $68.20\pm0.92$ & $83.63\pm0.62$ \\ 
20 & Aspect-Ratio, Rotation, Scale & $68.07\pm0.90$ & $83.53\pm0.61$ \\ 
\hline 

\hline

\hline
\end{tabular}
\end{center}\vspace{-2mm}
\caption{Ablation Study on \textbf{CIFAR-FS} validation set with different values of $M$. We choose $M=16$ for all the experiments. }
\label{tab:m_ablation}
\end{table}








\begin{figure}[h]
\begin{center}
  \includegraphics[width=1.0\linewidth]{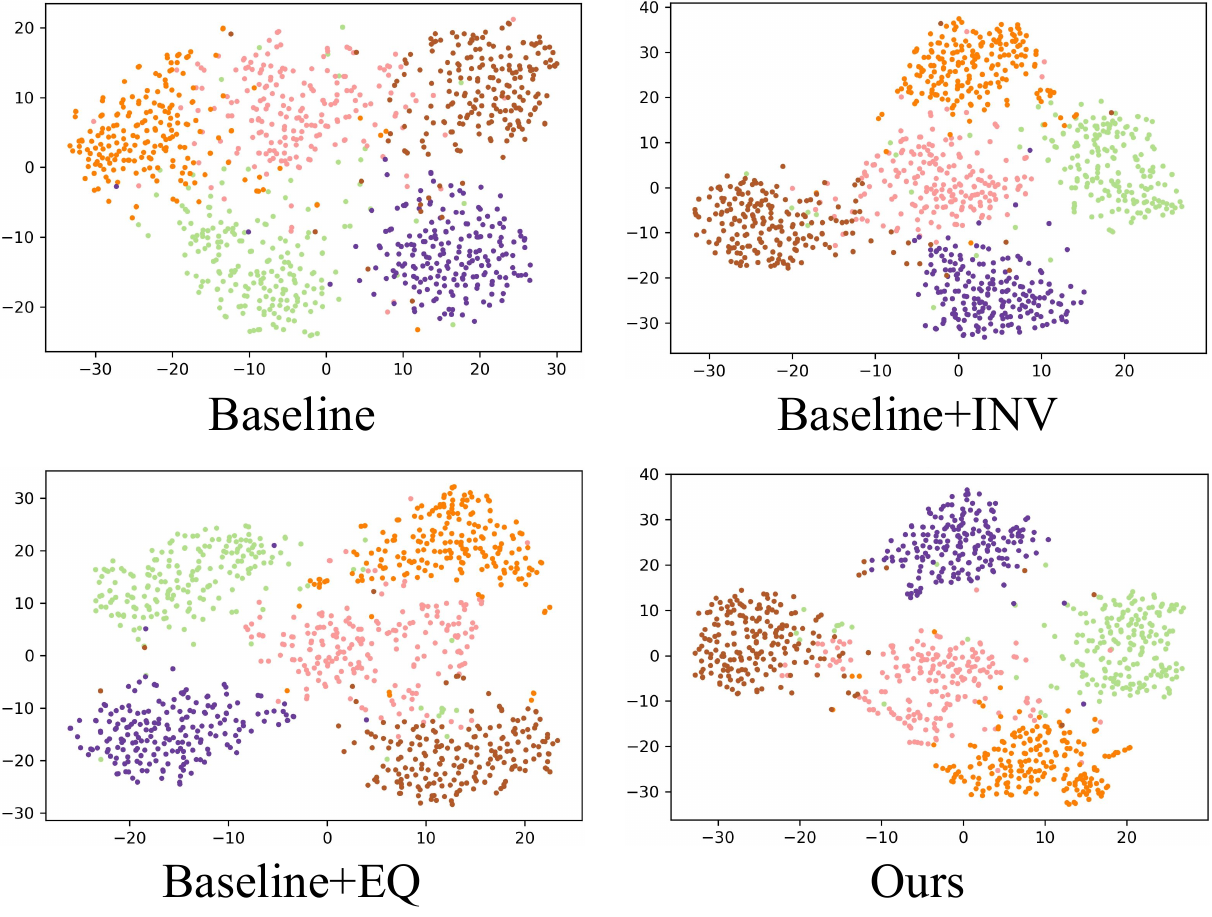}
\end{center}
\vspace{-4mm}
  \caption{t-SNE visualization of features for 1000 randomly sampled images from 5 randomly selected test classes of \textbf{miniImageNet} dataset. In our case, the learned embeddings provide better discrimination for unseen test classes.}
  \label{fig:tsne}
  \vspace{-5mm}

\end{figure}

\subsection{Analysis}

We do a t-SNE visualization of the output embeddings from $f_\Theta$ for the test images of miniImageNet to demonstrate the effectiveness of our method (see Fig.~\ref{fig:tsne}). We observe that the base learner trained in a supervised manner can retain good class discrimination even for unseen test classes. However, as evident in Fig.~\ref{fig:tsne}, the class boundaries are not precise and compact. Enforcing invariance on top of the base learner leads to more compact class boundaries; however, the sample embeddings of different classes are still relatively closer to one another. On the other hand, enforcing equivariance leads to class representations that are well spread out since it retains the transformation equivariant information in the embedding space. Finally, our proposed method takes advantage of both of these complementary properties and generates embeddings that lead to more compact clusters and discriminative class boundaries.

\subsection{Alternate Self-Supervision Losses}

In Table~\ref{tab:ssl_ablation}, to further analyze the performance improvement of our method, we conduct a set of experiments where commonly used self-supervised objectives like solving jigsaw puzzles \cite{noroozi2016unsupervised}, patch location prediction \cite{sun2019unsupervised}, context prediction \cite{doersch2015unsupervised}, rotation classification \cite{gidaris2018unsupervised} are added on top of the base learner as an auxiliary task. We found that our proposed method which aims to learn representations that retain both transformation invariant and equivariant information outperforms all of these SSL tasks by a good margin. Besides, we have noticed that the patch-based SSL tasks \cite{noroozi2016unsupervised, sun2019unsupervised, doersch2015unsupervised} generally underperform in comparison to SSL tasks that rely on changing the global statistics of the image while maintaining the local statistics; this conclusion is in line with the experimental results from \cite{gidaris2019boosting}. 

\begin{table}[t]
\begin{center}
\small
\resizebox{\columnwidth}{!}{%
\begin{tabular}{@{\hskip0pt}c@{\hskip6pt}c@{\hskip8pt}c@{\hskip1pt}}
\hline

\hline

\hline\\[-3mm]
 \textbf{Method}  & \textbf{1-Shot} & \textbf{5-Shot} 
 \\[-3mm]
\\
 \hline

\hline

\hline
Baseline Training & $62.02\pm0.63$ & $79.64\pm0.44$ \\
Baseline + Jigsaw Puzzle \cite{noroozi2016unsupervised} & $63.98\pm0.79$ & $81.08\pm0.55$ \\
Baseline + Location Pred \cite{sun2019unsupervised} & $64.39\pm0.81$ & $81.75\pm0.54$ \\
Baseline + Context Pred \cite{doersch2015unsupervised} & $64.72\pm0.79$ & $81.83\pm0.54$ \\
Baseline + Rotation \cite{gidaris2018unsupervised} & $65.25\pm0.80$ & $82.85\pm0.54$ \\
Ours (W/O KD) & $66.82\pm 0.80$ & $84.35\pm 0.51$ \\ \hline

\hline

\hline
\end{tabular}}
\end{center}
\vspace{-2mm}
\caption{FSL with different SSL objectives on \textbf{miniImageNet} dataset.}
\label{tab:ssl_ablation}
\vspace{-5mm}
\end{table}


\section{Conclusion}
In this work, we explored a set of inductive biases that help us learn highly discriminative and transferable representations for FSL. Specifically, we showed that simultaneously learning equivariant and invariant representations to a set of generic transformations results in retaining a complimentary set of features that work well for novel classes. We also designed a novel multi-head knowledge distillation objective which delivers additional gains. We conducted extensive ablation to empirically validate our claim that joint optimization for invariance and equivariance leads to more generic and transferable features. We obtained new state-of-the-art results on four popular benchmark FSL datasets as well as on the newly proposed challenging Meta-Dataset.



\vspace{-3mm}
{\small
\paragraph{Acknowledgements}
This research is based upon work supported by the Office of the Director of National Intelligence (ODNI), Intelligence Advanced Research Projects Activity (IARPA), via IARPA R\&D Contract No. D17PC00345. The views and conclusions contained herein are those of the authors and should not be interpreted as necessarily representing the official policies or endorsements, either expressed or implied, of the ODNI, IARPA, or the U.S. Government. The U.S. Government is authorized to reproduce and distribute reprints for Governmental purposes notwithstanding any copyright annotation thereon.}




\appendix
\section{Supplementary  Materials Overview}
In the supplementary materials we include the following: additional details about the applied geometric transformations (Section \ref{sec:geom_transformation}), additional results with the transformations sampled from the complete space of affine transformations (Section \ref{sec:affine_transformation}), ablation study on the coefficient of inductive loss (Section \ref{sec:loss_coeff}), ablation study on the temperature of knowledge distillation (Section \ref{sec:temp_KD}), effect of successive self knowledge distillation (Section \ref{sec:self_KD}), and effect of enforcing invariance and equivariance for supervised classification (Section \ref{sec:supervised_exp}).

\section{Geometric Transformations}
\label{sec:geom_transformation}
For our geometric transformations, we sample from a complete space of similarity transformation and use four rotation transformations: \{0\degree, 90\degree, 180\degree, 270\degree\}, two scaling transformations: \{0.67, 1.0\} and three aspect ratio transformations: \{0.67, 1.0, 1.33\}. Different combinations of these transformations lead to different values of $M$ (total number of applied transformations). An ablation study on the value of $M$ is included in section 4.2 of the main paper. In Table \ref{tab:m_description} we include the complete description of different values of $M$ that we use in our experiments. 

\begin{table*}
\begin{center}
\small
\begin{tabular}{cc}
\hline

\hline

\hline\\[-3mm]
 $M$ & \textbf{Description}  
 \\[-3mm]
\\
 \hline

\hline

\hline
3 & \textbf{AR}:\{0.67, 1.0, 1.33\} \\
4 & \textbf{ROT}:\{0\degree, 90\degree, 180\degree, 270\degree\}  \\ 
8 & \textbf{ROT}:\{0\degree, 90\degree, 180\degree, 270\degree\}\Cross\textbf{S}:\{0.67, 1.0\} \\ 
12 & \textbf{AR}:\{0.67, 1.0, 1.33\}\Cross\textbf{ROT}:\{0\degree, 90\degree, 180\degree, 270\degree\}  \\

16 & \Big(\textbf{AR}:\{0.67, 1.0, 1.33\}\Cross\textbf{ROT}:\{0\degree, 90\degree, 180\degree, 270\degree\}\Big) $\bigcup$ \Big(\textbf{ROT}:\{0\degree, 90\degree, 180\degree, 270\degree\}\Cross\textbf{S}:\{0.67\}\Big) \\

20 & \Big(\textbf{AR}:\{0.67, 1.0, 1.33\}\Cross\textbf{ROT}:\{0\degree, 90\degree, 180\degree, 270\degree\}\Big) $\bigcup$ \Big(\textbf{ROT}:\{0\degree, 90\degree, 180\degree, 270\degree\}\Cross\textbf{S}:\{0.67\}\Cross\textbf{AR}:\{0.67, 1.33\}\Big) \\ 
24 & \textbf{AR}:\{0.67, 1.0, 1.33\}\Cross\textbf{ROT}:\{0\degree, 90\degree, 180\degree, 270\degree\}\Cross\textbf{S}:\{0.67, 1.0\} \\
\hline 

\hline

\hline
\end{tabular}
\end{center}
\caption{Complete description of different values of $M$ based on different combination of aspect ratio (\textbf{AR}), rotation (\textbf{ROT}), and scaling (\textbf{S}) transformations.}
\label{tab:m_description}
\end{table*}

\section{Additonal Resutls with Affine Transformations}
\label{sec:affine_transformation}
We perform a set of experiments where the objective is to sample geometric transformation from the complete space of affine transformations. To this end, we quantize the affine transformation space according to Table \ref{tab:affine_quantized}. This leads to 972 distinct geometric transformations. Since it's not feasible to apply all the 972 transformations  on an input image $\mathbf{x}$ to obtain the input tensor $\mathbf{x}_{all} = \{\mathbf{x}_0, \mathbf{x}_1,...,\mathbf{x}_{971}\}$, we randomly sample 10 geometric transformations from the set of 972 transformations. We apply these randomly sampled 10 geometric transformations on an input image $\mathbf{x}$ and generate the input tensor $\mathbf{x}_{all}$. The results of these experiments are presented in Table \ref{tab:mini_affine}. From Table \ref{tab:mini_affine} it's evident that training with either invariance or equivariance improves over the baseline training for both 1 and 5 shot tasks (2.5-3.7\% improvement). Joint optimization for both invariance and equivariance provides additional improvement of $\sim$ 1\%. Even though the experiments with geometric transformations sampled from the complete affine transformation space do not improve over the training with $M=16$ (description of $M=16$ is available in Table \ref{tab:m_description}), the experiments demonstrate consistent improvement when both invariance and equivariance are enforced simultaneously. This provides additional support for our claim that enforcing both invariance and equivariance is beneficial for learning good general representations for solving challenging FSL tasks.

\begin{table}
\begin{center}
\small
\begin{tabular}{cc}
\hline

\hline

\hline\\[-3mm]
 \textbf{Transformation} & \textbf{Quantized Values}  
 \\[-3mm]
\\
 \hline

\hline

\hline
Rotation & \{0\degree, 90\degree, 180\degree, 270\degree \} \\
Translation(X) & \{$-$0.2, 0.0, 0.2 \}  \\ 
Translation(Y) & \{$-$0.2, 0.0, 0.2\}  \\
Scale & \{0.67, 1.0, 1.33\} \\
Aspect-Ratio & \{0.67, 1.0, 1.33\} \\
Shear & \{$-$20\degree, 0\degree, 20\degree\} \\
\hline 

\hline

\hline
\end{tabular}
\end{center}
\caption{Quantization of the space of Affine transformations.}
\label{tab:affine_quantized}
\end{table}

\begin{table}
\begin{center}
\small
\resizebox{\columnwidth}{!}{%
\begin{tabular}{@{\hskip0pt}ccc@{\hskip0pt}}
\hline

\hline

\hline\\[-3mm]
 \textbf{Method}  & \textbf{1-Shot} & \textbf{5-Shot} 
 \\[-3mm]
\\
 \hline

\hline

\hline
Baseline Training & $62.02\pm0.63$ & $79.64\pm0.44$ \\
Ours with only Invar (affine) & $65.55\pm0.81$ & $82.17\pm0.52$ \\
Ours with only Equi (affine) & $65.70\pm0.79$ & $82.47\pm0.53$ \\
Ours with Equi and Invar (affine) & $66.82\pm0.79$ & $82.96\pm0.53$ \\
Ours with Equi and Invar ($M$=16) & $66.82\pm0.80$ & $84.35\pm0.51$ \\ \hline

\hline

\hline
\end{tabular}}
\end{center}
\caption{Average 5-way few-shot classification accuracy with 95\%
confidence intervals on \textbf{miniImageNet} dataset; trained with different geometric transformations.}
\label{tab:mini_affine}
\end{table} 

\section{Ablation Study for Coefficient of Inductive Loss}
\label{sec:loss_coeff}

\begin{figure}
\begin{center}
  \includegraphics[width=1.0\linewidth]{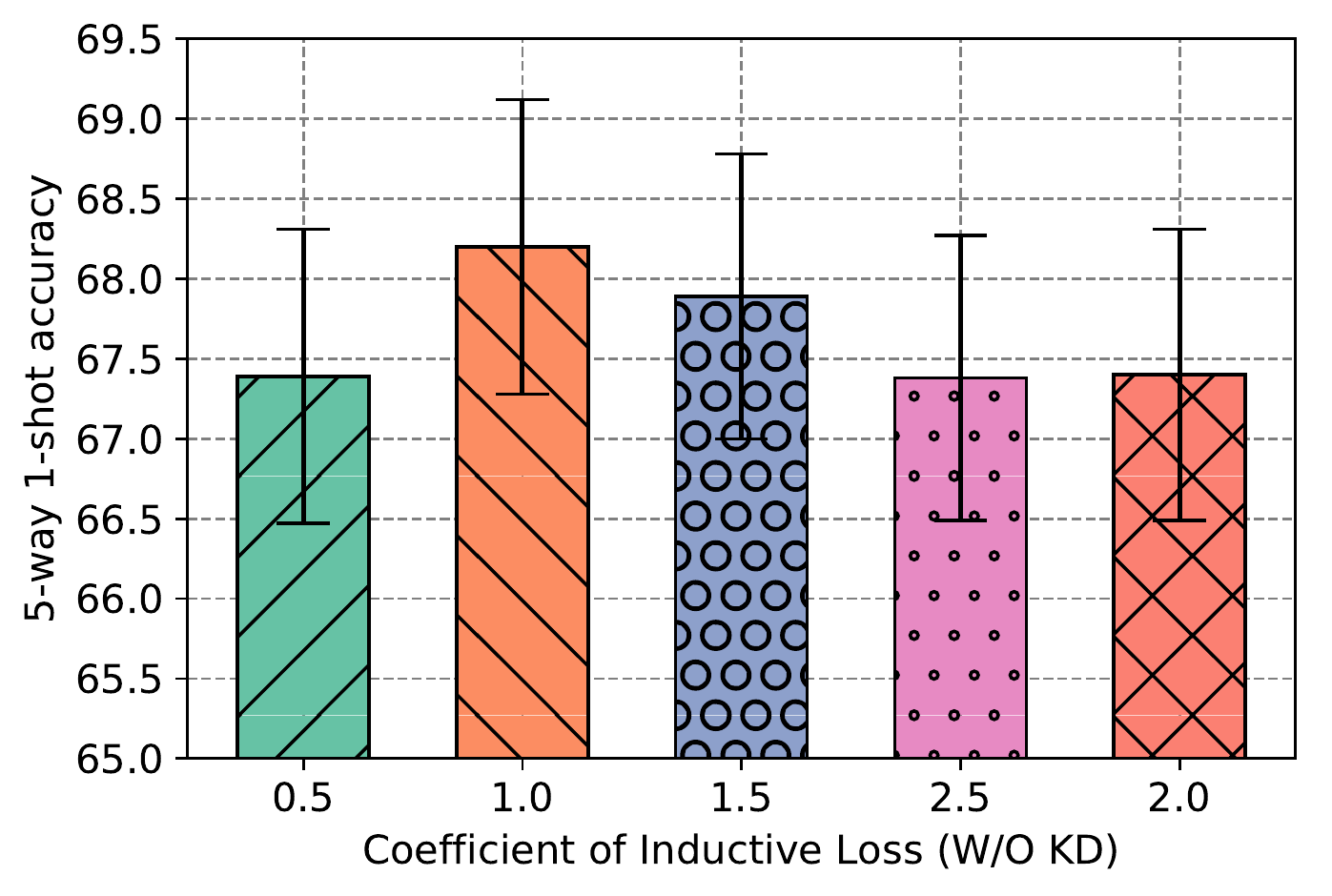}
\end{center}
\vspace{-4mm}
  \caption{Ablation study on \textbf{CIFAR-FS} validation set with different coefficients of the inductive loss (W/O KD); the reported score is average 5-way 1-shot classification accuracy with 95\% confidence intervals.}
  \label{fig:loss_coeff_1shot}

\end{figure}

We conduct an ablation study to measure the effect of different values of the coefficient of inductive loss (without multi-head distillation) on the CIFAR-FS \cite{bertinetto2018metalearning} validation set; the results of 5-way 1-shot FSL tasks are presented in fig. \ref{fig:loss_coeff_1shot}. From fig.\ref{fig:loss_coeff_1shot} it is evident that the proposed method is fairly robust to the different values of the coefficient of the inductive loss. However, the best performance is obtained when we set the loss coefficient to 1.0. Based on this ablation study, we use a loss coefficient of 1.0 for the inductive loss in all of our experiments.    

\section{Ablation Study for Knowledge Distillation Temperature}
\label{sec:temp_KD}

\begin{figure}
\vspace{-2mm}
\begin{center}
  \includegraphics[width=1.0\linewidth]{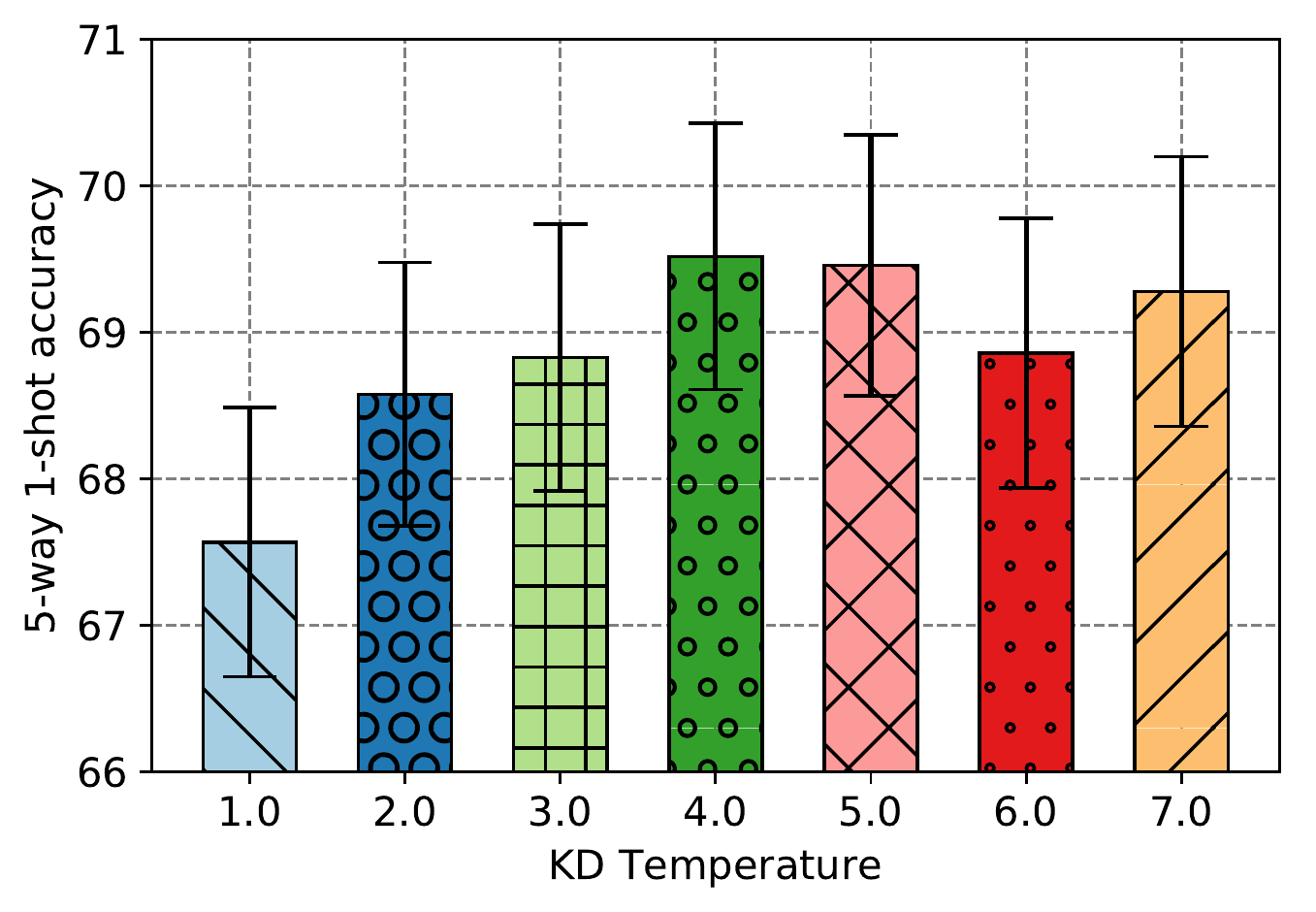}
\end{center}
\vspace{-4mm}
  \caption{Ablation study on \textbf{CIFAR-FS} validation set with different values of knowledge distillation temperature; the reported score is average 5-way 1-shot classification accuracy with 95\% confidence intervals.}
  \label{fig:temp_kd_1shot}

\end{figure}

To analyse the effect of knowledge distillation temperature (for Kullback Leibler (KL) divergence losses) we conduct an ablation study on the validation set of CIFAR-FS \cite{bertinetto2018metalearning} dataset. From fig. \ref{fig:temp_kd_1shot} we can observe that the proposed method with multi-head distillation objective is not very sensitive to the temperature coefficient of knowledge distillation. The proposed method achieves similar performance on the CIFAR-FS validation set when the value of distillation temperature is set to 4.0 and 5.0. Based on this ablation study and to be consistent with \cite{tian2020rethink}, we set the value of the coefficient of knowledge distillation temperature to 4.0 in all of our experiments.

\section{Effect of Successive Distillation}
\label{sec:self_KD}
\begin{figure}
\begin{center}
  \includegraphics[width=1.0\linewidth]{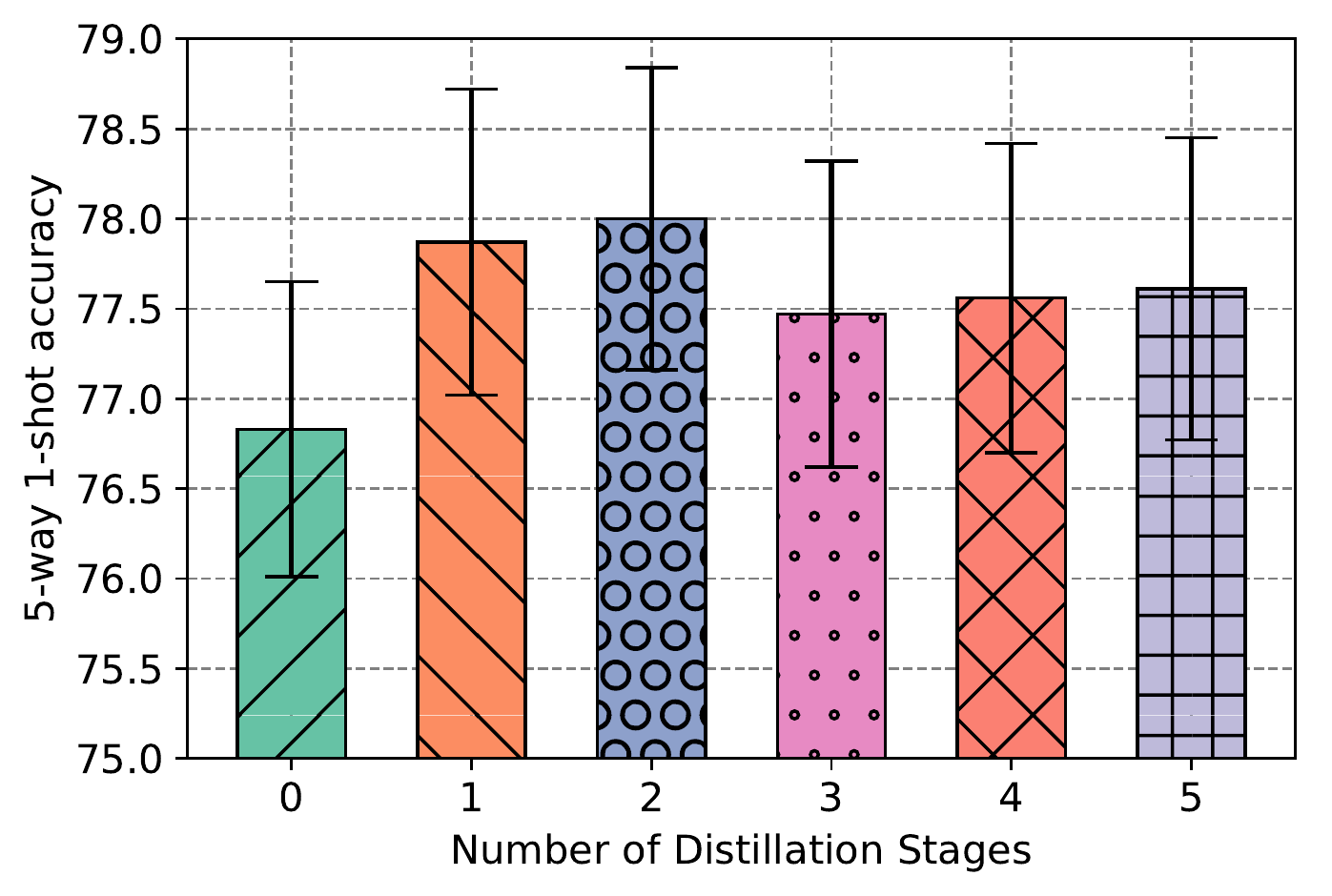}
\end{center}
\vspace{-4mm}
  \caption{Evaluation of different knowledge distillation stages on \textbf{CIFAR-FS} dataset; the reported score is average 5-way 1-shot classification accuracy with 95\% confidence intervals.}
  \label{fig:distill_1shot}

\end{figure}

In all of our experiments, we use only one stage of multi-head knowledge distillation. To further investigate the effect of knowledge distillation we perform multiple stages of self knowledge distillation on CIFAR-FS \cite{bertinetto2018metalearning} dataset. The results are presented in fig. \ref{fig:distill_1shot}. Here, the 0$^{th}$ distillation stage is the base learner trained with only the supervised baseline loss ($\mathcal{L}_{baseline}$), equivariant loss ($\mathcal{L}_{eq}$), and invariant loss ($\mathcal{L}_{in}$). From fig. \ref{fig:distill_1shot}, we observe that the performance in the FSL task improves for the first 2 stages of distillation, after that the performance saturates. Besides, the improvement from stage 1 to stage 2 is minimal ($\sim$ 0.1\%). Therefore, to make the proposed method more computationally efficient we perform only one stage of distillation in all of our experiments.

\section{Invariance and Equivariance for Supervised Classification}
\label{sec:supervised_exp}
To demonstrate the effectiveness of complementary strengths of invariant and equivariant representations we conduct fully supervised classification experiments on benchmark CIFAR-100 dataset \cite{CIFAR-100}. For these experiments, we use the standard Wide-Resnet-28-10 \cite{BMVC2016_87} architecture as the backbone. For training, we use an SGD optimizer with an initial learning rate of 0.1. We set the momentum to 0.9 and use a weight decay of 5e$-$4. For all the experiments, the training is performed for 200 epochs where the learning rate is decayed by a factor of 5 at epochs 60, 120, and 160. We use a batch size of 128 for all the experiments as well as a dropout rate of 0.3. The training augmentations include standard data augmentations: random crop and random horizontal flip. For enforcing invariance and equivariance, we set the value of $M$ to 12 for computational efficiency; description of $M=12$ is available in Table \ref{tab:m_description}. We do not perform knowledge distillation for these experiments. The results of these experiments are presented in Table \ref{tab:supervised}. 

From Table \ref{tab:supervised}, we can notice that enforcing invariance provides little improvement (0.2\%) over the supervised baseline. This is expected since the train and test data is coming from the same distribution and same set of classes; making the class boundaries compact (for seen classes) doesn't provide that much additional benefit. However, in the case of FSL we observe that enforcing invariance over baseline provides 2.62\%, 2\%, and 3.5\% improvement for miniImageNet \cite{NIPS2016_6385}, CIFAR-FS \cite{bertinetto2018metalearning}, and FC100 \cite{oreshkin2018tadam} datasets respectively (section 4.2 of main text). On the other hand, enforcing equivariance for supervised classification provides better improvement (1.8\%) since it helps the model to better learn the structure of data. Even though enforcing equivariance provides noticeable improvement for supervised classification, in the case of FSL we obtain a much bigger improvement of 4.07\%, 4.87\%, and 4.13\% for miniImageNet \cite{NIPS2016_6385}, CIFAR-FS \cite{bertinetto2018metalearning}, and FC100 \cite{oreshkin2018tadam} datasets respectively (section 4.2 of main text). Finally, joint optimization for both invariance and equivariance achieves the best performance and provides minimal but consistent improvement of 0.1\% over enforcing only equivariance. However, joint optimization provides a much larger improvement on FSL tasks (see section 4.2 of the main text). From these experiments, we conclude that, although enforcing both invariance and equivariance is beneficial for supervised classification, injecting these inductive biases is more crucial for FSL tasks since the inductive inference for FSL tasks is more challenging (inference on unseen/novel classes).



\begin{table}
\begin{center}
\small
\begin{tabular}{cc}
\hline

\hline

\hline\\[-3mm]
\textbf{Method}  & \textbf{Error Rate (\%)} 
 \\[-3mm]
\\
 \hline

\hline

\hline

Supervised Baseline & $18.78$ \\
Ours with only Invariance & $18.56$ \\ 
Ours with only Equivariance & $16.95$ \\ 
Ours with Equi and Invar (W/O KD) & $16.84$ \\
\hline 

\hline

\hline
\end{tabular}
\end{center}
\caption{Results with invariance and equivariance for supervised classification on \textbf{CIFAR-100} dataset.}
\label{tab:supervised}
\end{table}



{\small
\bibliographystyle{cvpr}
\bibliography{cvpr}
}

\end{document}